\documentclass[10pt,twocolumn,letterpaper]{article}

\usepackage{cvpr}              

\usepackage[accsupp]{axessibility}
\usepackage{graphicx}
\usepackage{amsmath}
\usepackage{amssymb}
\usepackage{booktabs}
\usepackage{float}
\usepackage[usenames,dvipsnames]{xcolor}
\usepackage[pdftex,outline]{contour}
\usepackage{tabularx}
\usepackage{overpic}
\usepackage{algpseudocode}
\usepackage{algorithm}

\newif\ifdraft
\draftfalse

\ifdraft
\newcommand{\vk}[1]{{\color{purple}[\textbf{Vova:} #1]}}
\newcommand{\rhc}[1]{{\color{blue}[\textbf{Rana:} #1]}}
\newcommand{\noam}[1]{{\color{red}[\textbf{Noam:} #1]}}
\newcommand{\rl}[1]{{\color{green}[\textbf{Richard:} #1]}}

\newcommand\rsout{\bgroup\markoverwith{\textcolor{red}{\rule[0.5ex]{2pt}{0.4pt}}}\ULon}
\else
\newcommand{\rhc}[1]{}
\newcommand{\vk}[1]{}
\newcommand{\rl}[1]{}
\newcommand{\noam}[1]{}
\newcommand\rsout{}

\fi

\newif\ifreviewer
\reviewertrue
\reviewerfalse
\ifreviewer
\newcommand{\rev}[1]{{\color{black}#1}}
\else
\newcommand{\rev}[1]{}
\fi

\newcommand{\ourmethod}{DA Wand}

\newcommand{\M}{\mathbf{M}}
\newcommand{\F}{\mathbf{F}}
\newcommand{\V}{\mathbf{V}}
\newcommand{\tri}{\mathbf{t}}
\newcommand{\seg}{\mathbf{\bar{F}}}\newcommand{\U}{\mathbf{{U}}}

\newtheorem{theorem}{Theorem}

\usepackage[pagebackref,breaklinks,colorlinks]{hyperref}

\usepackage[capitalize]{cleveref}
\crefname{section}{Sec.}{Secs.}
\Crefname{section}{Section}{Sections}
\Crefname{table}{Table}{Tables}
\crefname{table}{Tab.}{Tabs.}

\def\cvprPaperID{8792} 
\def\confName{CVPR}
\def\confYear{2023}

\begin{document}
\title{DA Wand: Distortion-Aware Selection using Neural Mesh Parameterization}

\author{\textbf{Richard Liu}\\
University of Chicago\\
\and
\textbf{Noam Aigerman}\\
Adobe Research\\
\and
\textbf{Vladimir G. Kim}\\
Adobe Research\\
\and
\textbf{Rana Hanocka}\\
University of Chicago\\
}
\twocolumn[{%
\renewcommand\twocolumn[1][]{#1}%
\maketitle
\begin{center}
    \vspace*{-0.3in}
    \centering
    \captionsetup{type=figure}
    \includegraphics[width=\textwidth]{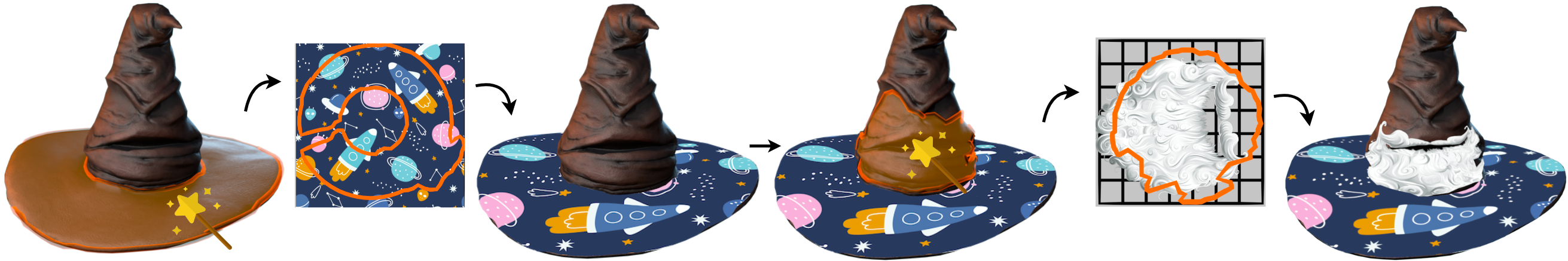} 
    \vspace{-0.4cm}
    \caption{\ourmethod{} enables interactive decaling through a conditional selection of local \textit{distortion-aware} patches. 
    Our method identifies large patches within both developable and high-curvature regions which induce a low distortion parameterization.}
    \label{fig:teaser}
\end{center}%
}]

\begin{abstract}
We present a neural technique for learning to select a local sub-region around a point which can be used for mesh parameterization. The motivation for our framework is driven by interactive workflows used for decaling, texturing, or painting on surfaces. Our key idea is to incorporate segmentation probabilities as weights of a classical parameterization method, implemented as a novel differentiable parameterization layer within a neural network framework. We train a segmentation network to select 3D regions that are parameterized into 2D and penalized by the resulting distortion, giving rise to segmentations which are distortion-aware. Following training, a user can use our system to interactively select a point on the mesh and obtain a large, meaningful region around the selection which induces a low-distortion parameterization. Our code\footnote{\href{https://github.com/threedle/DA-Wand}{https://github.com/threedle/DA-Wand}} and project\footnote{\href{https://threedle.github.io/DA-Wand/}{https://threedle.github.io/DA-Wand/}} are publicly available. 
\end{abstract}

\vspace{-5mm}
\section{Introduction}
Many interactive workflows for decaling, texturing, or painting on a 3D mesh require extracting a large surface patch around a point that can be mapped to the 2D plane with low distortion. Unlike \emph{global} parameterization approaches that map the entire mesh to 2D while introducing as few cuts as possible~\cite{desbrun_intrinsic_2002, mullen_spectral_2008,floater_mean_2003,levy_least_2002,sawhney_boundary_2018,hormann_mesh_2008,rabinovich_discrete_2017,smith_bijective_2015,aigerman_orbifold_2015,sheffer_abf_2005,claici_isometry-aware_2017,sander_texture_2001,shtengel_geometric_2017}), this work focuses on segmenting a \emph{local sub-region} around a point of interest on a mesh for parameterization ~\cite{schmidt_interactive_2006,zhang_deep_2020,sun_texture_2013,schmidt_stroke_2013,xu_texture_2014,melvaer_geodesic_2012,xu_field-aware_2019, kokkinos_intrinsic_2012}. Local parameterizations are advantageous in certain modeling settings because they are inherently user-interactive, can achieve lower distortion than their global counterparts, and are computationally more efficient. To date, however, techniques for extracting a surface patch that is amenable to local parameterization have largely relied on heuristics balancing various compactness, patch size, and developability priors~\cite{julius_d-charts_2005}. 

This work instead takes a data-driven approach to learn \emph{distortion-aware} local segmentations that are optimal for local parameterization. Our proposed framework uses a novel \emph{differentiable parameterization layer} to predict a patch around a point and its corresponding UV map. This enables \emph{self-supervised} training, in which our network is encouraged to predict area-maximizing and distortion-minimizing patches through a series of carefully constructed priors, allowing us to sidestep the scarcity of parameterization-labeled datasets.

We name our system the \textbf{Distortion-Aware Wand (DA Wand)}, which given an input mesh and initial triangle selection, outputs soft segmentation probabilities. We incorporate these probabilities into our parameterization layer by devising a weighted version of a classical parameterization method - \emph{LSCM}~\cite{levy_least_2002} -  which we call \emph{wLSCM}. This adaptation gives rise to a \emph{probability-guided parameterization}, over which a distortion energy can be computed to enable \emph{self-supervised} training. We prove a theorem stating that the wLSCM UV map converges to the LSCM UV map as the soft probabilities converge to a binary segmentation mask, which establishes the direct relation between probabilities and binary segmentation in the parameterization context.

Reducing the distortion of the UV map and maximizing the segmentation area are competing objectives, as UV distortion scales monotonically with patch size. Naively summing these objectives leads to poor optimization with undesirable local minima. We harmonize these objectives by devising a novel \emph{thresholded-distortion loss}, which penalizes triangles with distortion above some user-prescribed threshold. We additionally encourage compactness through a \emph{smoothness loss} inspired by the graphcuts algorithm \cite{kolmogorov_what_2004}.


We design a novel near-developable segmentation dataset to initialize the weights of our segmentation network, with an automatic generation algorithm which can be run out of the box. We then train this network end-to-end on a dataset of unlabelled natural shapes using our parameterization layer with distortion and compactness priors. 


We leverage a MeshCNN~\cite{hanocka_meshcnn_2019} backbone to learn directly on the input triangulation which enables sensitivity to sharp features and a large receptive field which enables patch growth. Moreover, by utilizing \textit{intrinsic} mesh features as input, our system remains invariant to rigid-transformations. 

\ourmethod{} allows a user to interactively select a triangle on the mesh and obtain a large, meaningful region around the selection which can be UV parameterized with low distortion. We show that the neural network is able to leverage global information to extend the segmentation with minimal distortion gain, in contrast to existing heuristic methods which stop at the boundaries of high curvature regions. Our method can produce user-conditioned segmentations at interactive rates, beating out alternative techniques. We demonstrate a compelling interactive application of \ourmethod{} in \cref{fig:teaser}, in which different regions on the sorting hat mesh are iteratively selected and decaled. We show additional example textures in the supplemental. 



\section{Related Work}
\label{sec:relatedwork}
\subsection{Global Cutting \& Parameterization}
Surface parameterization has been a long-standing geometry processing problem. An extensive line of work has dealt with \textit{global} surface parameterization, in which a 3-dimensional surface, discretized as a triangle mesh with fixed connectivity, is mapped to the plane \cite{desbrun_intrinsic_2002, mullen_spectral_2008,floater_mean_2003,levy_least_2002,sawhney_boundary_2018,hormann_mesh_2008,rabinovich_discrete_2017,smith_bijective_2015,aigerman_orbifold_2015,sheffer_abf_2005,claici_isometry-aware_2017,sander_texture_2001,shtengel_geometric_2017}). Most of these methods assume a disk topology surface, and some generate seams on a heuristic basis to ensure this. A few seamless approaches exist that instead introduce point singularities.  \cite{kharevych_discrete_2006,springborn_conformal_2008,myles_global_2012,gillespie_discrete_2021,li_computing_2022}.

Though most parameterization works assume a disk topology input, the target domain for many texturing and fabrication applications is often a closed mesh without boundaries, and segmenting a mesh into topological disks in an optimal way is a non-trivial problem. Thus, many works have focused on the sub-problem of cutting a closed surface into disks with desirable properties for the target application -- typically developability and minimal-length seams \cite{sheffer_seamster_2002,levy_least_2002,julius_d-charts_2005,gu_geometry_2002, lucquin_seamcut_2017,sharp_variational_2018,yamauchi_mesh_2005,wang_computing_2008,shatz_paper_2006,mitani_making_2004}. Some works jointly optimize for the parameterization and cutting objectives \cite{poranne_autocuts_2017,li_optcuts_2019}, taking advantage of the close relationship between seam placement and parameterization distortion. 

\subsection{Local Cutting \& Parameterization}
A few past works have examined the \emph{local} problem of segmenting and parameterizing a surface patch which encompasses a point of interest. Exponential Maps (more accurately referred by its inverse \emph{logarithmic map}, which is how we will refer to this method moving forward) is a cornerstone work in this domain \cite{schmidt_interactive_2006}, and subsequent local parameterization works have primarily extended the insights of the original method \cite{zhang_deep_2020,sun_texture_2013,suzuki_autocomplete_2017,schmidt_stroke_2013,xu_texture_2014,melvaer_geodesic_2012,xu_field-aware_2019}. All of the local parameterization works to date either do not rely on a local segmentation of the domain, grow out a fixed-radius geodesic circle, or apply some form of local shape analysis to iteratively floodfill the selection. This last class of chart growing techniques based on surface analysis is also performed by some global mesh decomposition methods, such as DCharts \cite{julius_d-charts_2005}, which makes these methods amenable for adaptation to the local segmentation setting. 

\begin{figure}[t]
    \centering
    \newcommand{\pl}{-3.5}
    \newcommand{\pt}{44.8}
    \begin{overpic}[width=0.7\columnwidth]{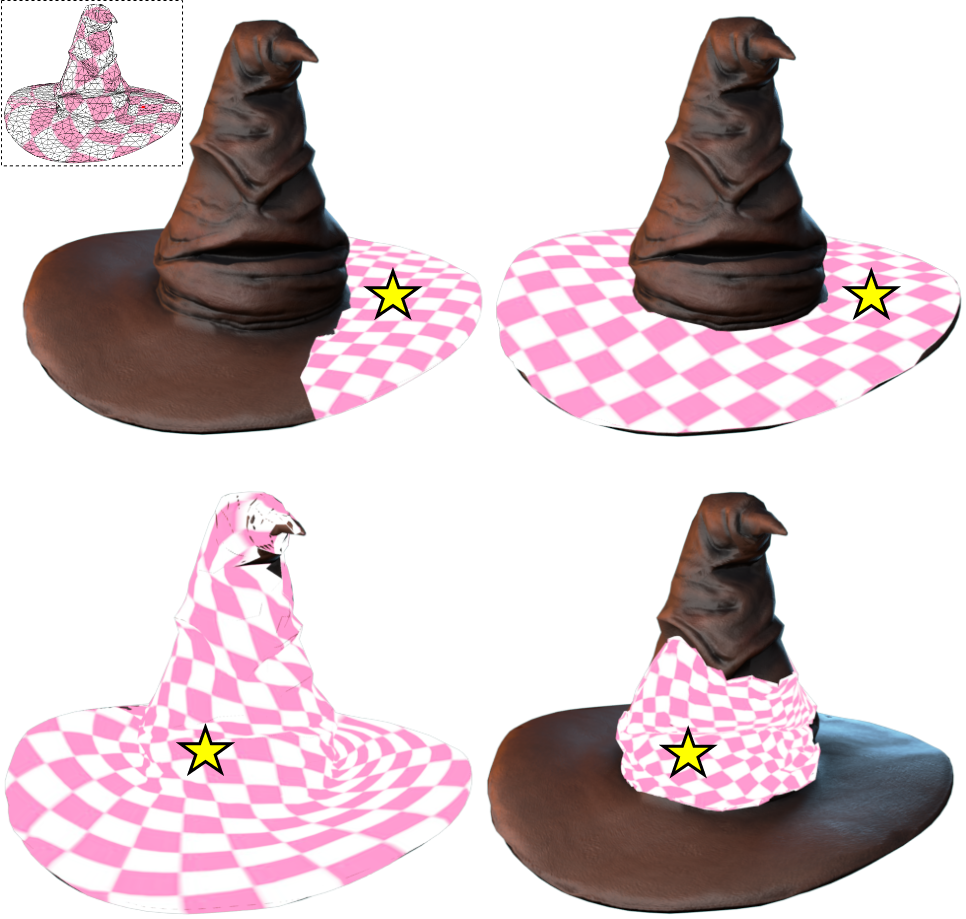}
    \put(7, \pt){\small\textcolor{black}{\textbf{Logarithmic Map}}}
    \put(14, \pl){\small\textcolor{black}{\textbf{DCharts}}}
    \put(60, \pl){\small\textcolor{black}{\textbf{Our Method}}}
    \put(60, \pt){\small\textcolor{black}{\textbf{Our Method}}}
    \end{overpic}
    \caption{Segmentations from our baseline results conditioned on the same selections (stars) as our method in \cref{fig:teaser}, with resulting UV textures. To produce the Logarithmic Map segmentation, we select a low-distortion patch from the full map (dashed inset).}
    \label{fig:teasercomparison}
\end{figure}

Our work is primarily interested in this \emph{local segmentation for parameterization} problem. To our knowledge, however, our work is the first to (1) make explicit the objective of a \emph{minimal-distortion}, \emph{large} local segmentation and (2) apply data-driven techniques to solve this problem. Our approach overcomes the limitations of the aforementioned heuristic methods, which preclude their use in certain texturing applications, shown in \cref{fig:teasercomparison}. 

\begin{figure*}
    \centering
    \includegraphics[width=\textwidth]{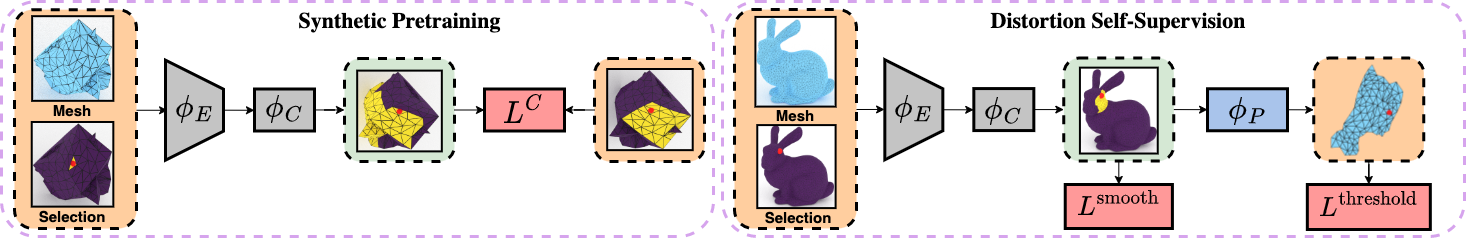}
    \vspace{-.6cm}
    \caption{\textbf{Overview.} The encoder $\phi_E$ maps a \textcolor{Apricot}{mesh} and a \textcolor{Apricot}{selection triangle} to edge features which are passed into $\phi_C$ to output a local \textcolor{YellowGreen}{segmentation}. The training is split into two phases. We have a \textbf{pretrain phase} where we jointly train our encoder and classifier network on a synthetic dataset of shapes with ground truth segmentations and randomly sampled initial selections. We supervise with cross entropy loss $\textcolor{red}{L^{C}}$. In our main \textbf{distortion self-supervision phase}, we finetune $\phi_E$ and $\phi_C$ on a dataset of natural shapes, where \emph{no ground truth segmentations exist}. We introduce a differentiable parameterization layer $\textcolor{RoyalBlue}{\phi_P}$ which computes a \textcolor{RoyalBlue}{weighted UV embedding} for the network-predicted segmentation on the fly during training. The UV map is sent into a distortion function of choice $D$, which is used in the thresholded-distortion loss $\textcolor{red}{L^{\text{threshold}}}$. The smoothness loss $\textcolor{red}{L^{\text{smooth}}}$ enforces a compactness prior on the network prediction, which helps curtail non-contiguous segmentations.}
    \label{fig:overview}
\end{figure*}

\subsection{Mesh Segmentation}
Our work also has a foot in the space of general mesh segmentation techniques, in which a mesh is partitioned into disjoint parts according to some set of criteria. One prominent class of segmentation techniques involves \emph{semantic segmentation}, where a mesh is decomposed into meaningful parts \cite{lai_feature_2006,lahav_meshwalker_2020,lai_rapid_2009,attene_hierarchical_2006,zhang_interactive_2011,van_kaick_prior_2011,kalogerakis_learning_2010,wang_projective_2013,xie_3d_2014,guo_3d_2015,sidi_unsupervised_2011,xu_style-content_2010,golovinskiy_consistent_2009,bergamasco_graph-based_2012,yan_variational_2012,zhang_shape_2015}. Another prominent class tackles the disk topology objective discussed in the previous section \cite{sheffer_seamster_2002,levy_least_2002,julius_d-charts_2005,gu_geometry_2002},\cite{lucquin_seamcut_2017,sharp_variational_2018,yamauchi_mesh_2005,wang_computing_2008,shatz_paper_2006,mitani_making_2004}. A closely related set of methods perform sharp feature identification, where the surface geometry is analyzed to identify boundaries of minimal principal curvature, and the mesh is segmented along these boundaries \cite{ghosh_fast_2013,juttler_isogeometric_2014,kim_new_2005,lien_approximate_2004,kaick_shape_2015,bajaj_convex_1992,asafi_weak_2013,zhuang_anisotropic_2014,kin-chung_au_mesh_2012,yan_variational_2012,wang_spectral_2014}. Recently, learning-based methods have been introduced for each of these objectives \cite{he_deep_2021,lahav_meshwalker_2020,kalogerakis_learning_2010,xie_3d_2014,guo_3d_2015,hanocka_meshcnn_2019,sharp_diffusionnet_2022,xu_directionally_2017}, but to our knowledge there does not exist a learning-based work that targets our objective of local distortion-aware segmentation.

\section{Method}
\label{sec:method}
We assume as input a triangle mesh $\M$ with faces $\F$ and vertices $\V$, and a  selected  triangle $\tri^*$.  Our goal is to define a binary segmentation of the mesh faces $\seg$ which contains $\tri^*$. Additionally, $\seg$ needs to satisfy three competing objectives: i) enclose a region that can be UV-mapped with low distortion; ii) cover an as-large-as-possible mesh area; iii) have a compact boundary, which we define as a boundary enclosing a single connected region on the mesh surface. We denote by $\U\in\mathbb{R}^{|\V|\times2}$ the computed UV mapping assigning a 2D coordinate to each vertex. 


Our framework predicts a \emph{soft} segmentation, in which each triangle is assigned a probability designating whether it belongs to the segmented region. The key contribution that drives our framework is a \emph{differentiable parameterization layer} (Sec~\ref{subsec:classifier_driven_param}), which computes a UV parameterization conditioned on the segmentation probabilities. To train this architecture we introduce novel distortion and compactness objectives favoring large contiguous segments with low-distortion UV parameterizations (Sec~\ref{subsec:objectives}). 


Training is conducted in two stages: i) in pre-training, we leverage a novel synthetic dataset (Sec~\ref{subsec:synth_data}), with a ground truth decomposition into near-developable regions, and train our segmentation network with strong supervision. ii) we then train the entire system end-to-end with distortion self-supervision (enabled by the parameterization layer) on a dataset of unlabelled natural shapes (Sec ~\ref{subsec:training}). An overview of the system is shown in \cref{fig:overview}.




\subsection{Probability-Guided Parameterization}
\label{subsec:classifier_driven_param}
Our goal with the parameterization layer is to devise a fast, differentiable mapping from the segmentation $W$ to a low-distortion UV map. To predict the segmentation, we leverage MeshCNN, a robust mesh segmentation  architecture~\cite{hanocka_meshcnn_2019}. 
The network predicts a soft segmentation $W$ which assigns a weight $w_\tri$ to each triangle $\tri$. This soft segmentation can be clamped to a binary segmentation (e.g. through rounding) to define a subregion $\seg$ of the mesh. The challenge is then, how to devise a parameterization that accounts for the soft segmentation and smoothly approximates the parameterization of the subregion $\seg$?

Our main observation is that we can consider a classic parameterization technique - Least Square Conformal Maps (LSCM)~\cite{levy_least_2002} - and define a \emph{weighted} version of it, dubbed \emph{wLSCM}, such that as the weights approach binary values $\left\{0,1\right\}$ which define a hard segmentation $\seg$, wLSCM converges to the (non-weighted) LSCM parameterization of $\seg$. We achieve this such that wLSCM can still be implemented as a straightforward differentiable layer.

The original LSCM method considers the vertices of each triangle $(v_1,v_2,v_3)$, and their image under the (unknown) UV map, $(u_1,u_2,u_3)$, and considers the affine transformation representing that map, which satisfies
\begin{equation}
    A_\tri v_i + \delta_\tri = u_i,\  i\in\left\{1,2,3\right\},
\end{equation}
where $A_t\in\mathbb{R}^{2\times2}$, $\delta_t\in\mathbb{R}^2$ are the linear and translational components of the map, respectively (see~\cite{levy_least_2002}).

\begin{figure}[h]
    \centering
    \newcommand{\pl}{-3.5}
    \begin{overpic}[width=\columnwidth]{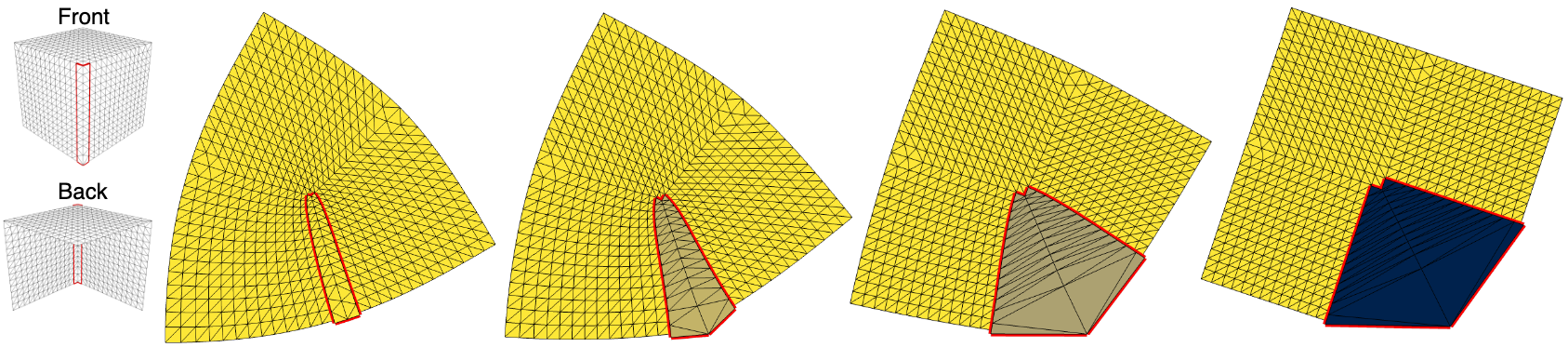}
    \put(14,  \pl){\small\textcolor{black}{$\mathbf{W = 1}$}}
    \put(34,  \pl){\small\textcolor{black}{$\mathbf{W = 0.3}$}}
    \put(57,  \pl){\small\textcolor{black}{$\mathbf{W = 0.2}$}}
    \put(81,  \pl){\small\textcolor{black}{$\mathbf{W = 0}$}}
    \end{overpic}
    \caption{We show the wLSCM UV map for a mesh made from 3 sides of a cube, with decreasing weights \textbf{W} for the region outlined in red. The wLSCM UV map of the non-outlined region when \textbf{W=0} is exactly the UV map of unweighted LSCM over the same region, as proved in \cref{thm:wlscm}.}
    \label{fig:lscmweights}
\end{figure}

LSCM then aims to minimize \emph{conformal error}, which measures the change of the triangle angles resulting from the map. It considers the closest similarity matrix $S(A_\tri)$ (a matrix with no conformal error), and measures the least-squares error between $A_t$ and the similarity matrix:

\begin{equation}
\label{e:lscm}
    E_\text{LSCM}=\sum_{\tri\in\F}||A_\tri - S(A_\tri)||^2.
\end{equation}
LSCM computes the UV parameterization $\U$ which minimizes this energy, which reduces to solving a sparse linear system. We propose \emph{relaxing} the segmentation problem by adding a weighting term using the weights $W$ predicted by the segmentation network: 

\begin{equation}
\label{e:wlscm}
    E_\text{wLSCM} = \sum_{\tri\in\F}w_\tri||A_\tri - S(A_\tri)||^2.
\end{equation}
The minimizer of this energy can still be obtained via a linear solve. Hence, our differentiable parameterization layer consists of plugging the predicted soft segmentation $W$ into the wLSCM energy (\cref{e:wlscm}), and solving the linear system to get the UV parameterization $\U$. ~\cref{fig:lscmweights} illustrates how weights affect the parameterization. Importantly, as the weights converge to a binary segmentation, the resulting wLSCM UV parameterization converges to the LSCM parameterization of the segmented region:
\begin{theorem}\label{thm:wlscm}
Let $\seg\subset\F$ be a subset of triangles of the mesh, which comprises one connected component. Let $W$ be  non-negative weights assigned to the triangles s.t. the weights of $\seg$ are non-zero. Let $\U_W$ be the minimizer of \cref{e:wlscm} w.r.t $W$. Then $\U_W$, restricted to $\seg$, is a well-defined, continuous function of $W$. Furthermore, if the non-zero weights of $W$ are all equal to $1$, then $\U_W$ restricted to $\seg$ is exactly equal to the (non-weighted) LSCM parameterization of $\seg$.
\end{theorem}
The theorem has a straightforward proof via continuity of polynomial roots. See the appendix for the full proof.\\\\
\textbf{Segmentation network architecture.} For the soft segmentation prediction, we adapt a 12-layer (6 down, 6 up) MeshCNN~\cite{hanocka_meshcnn_2019} to define an edge encoder, $\phi_E$. Each layer has 3 residual convolution blocks with hidden dimension 16. This fully-convolutional backbone enables a large receptive field with relatively few parameters, while the graph-based convolutions allow for the fine-grained edge understanding necessary to segment along noisy geometric features. We do not incorporate pooling in the lower layers, as we find it oversmooths the learned features. $\phi_E$ takes as input batched pairs of meshes and point selections $(M, x)$, and computes dihedral angles, the heat kernel signature (HKS) \cite{sun_concise_2009}, and a one-hot vector of selection triangle incidence as edge features.

A second per-triangle classifier $\phi_C$ predicts the per-triangle weights $W$. It is implemented as a 3-layer multi-layer perceptron (MLP) with dropout and a final sigmoid activation layer to map scalars to soft probabilities.

\begin{figure*}[t!]
    \centering
    \includegraphics[width=\textwidth]{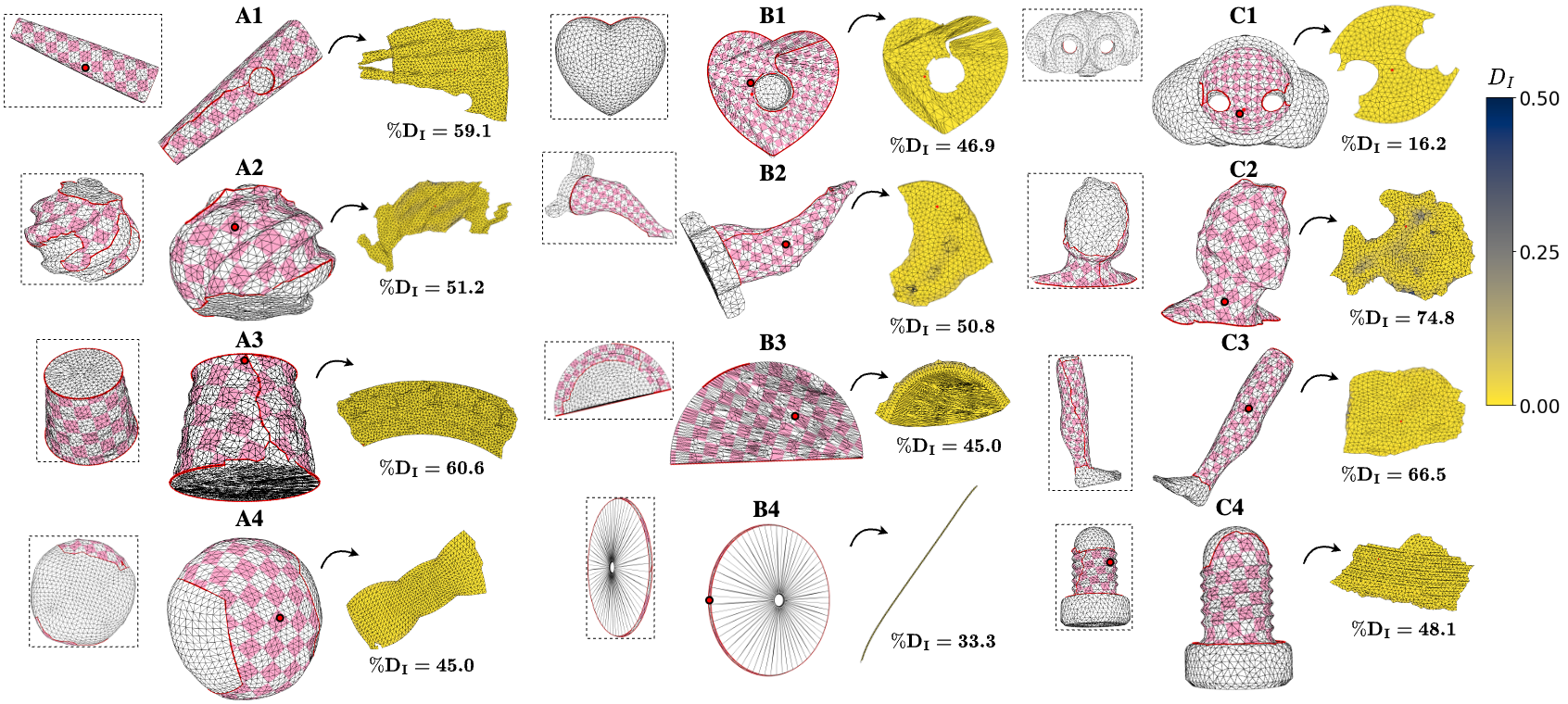}
    \vspace{-5mm}
    \caption{Conditional segmentations predicted by our method. We show the corresponding UVs computed with SLIM, colored by the isometric distortion metric $I$. Selection points are marked in \textcolor{red}{red}, and \textbf{\%}$\mathbf{D_I}$ is reported beneath each UV.}
    \label{fig:gallery}
\end{figure*}

\subsection{Distortion and Compactness Priors}
\label{subsec:objectives}
To train the network, we carefully design losses which balance across our three competing patch objectives of size, distortion, and compactness.\\\\
\textbf{Thresholded-distortion loss.} We aim to produce a UV parameterization with low isometric distortion, defined via the As-Rigid-As-Possible (ARAP) energy ($D_{arap}$) \cite{liu_localglobal_2008}.
\begin{equation}
    D_{\text{arap}}(x_t) = \sum_{i=0}^{2} \cot(\theta_t^i)||(x_t^i - x_t^{i+1}) - L_t(v_t^{i} - v_t^{i+1})||^2 
\end{equation}
where $v_t^{i}$ is the $i$-th vertex of triangle $t$ in local coordinates, $\theta_t^i$ is the angle opposite the edge ($x_t^i$, $x_t^{i+1}$), and $L_t$ is the ``best-fit" rotation matrix as described in \cite{liu_localglobal_2008}.

Typically the distortion of a UV parameterization grows monotonically with the size of the selection area. This puts in direct contrast our objectives of large patch area and low-distortion parameterization. We resolve this by formulating an objective that aims to maximize the the number of triangles with distortion under a \emph{threshold}. We write this objective in terms of a novel \emph{thresholded-distortion loss}:
\begin{equation}\label{eq:countloss}
    L^{\text{threshold}}(D) = \sum_t \frac{A_t}{\sum A_t} \left(1 - e^{-(D_t/\gamma)^\alpha}\right),
\end{equation}
where $A_t$ is the area of triangle $t$, $\gamma$ acts as a soft threshold on the per-triangle distortion $D_t$, and $\alpha$ is a hyperparameter which controls the sensitivity to the threshold. This loss acts as a weighted count of triangles with distortion below $\gamma$.
In all our experiments, we set $\gamma = 0.01$ and $\alpha = 5$. \\\\
%
\textbf{Compactness priors.} 
In order to achieve our last objective of a compact segmentation boundary, we introduce two priors into network training. First, we introduce a \emph{smoothness prior}, in the form of a \emph{smoothness loss} derived from the well-known graphcuts algorithm~\cite{kolmogorov_what_2004}. 
\begin{equation}\label{eq:smooth}
    L^{\text{smooth}} = \frac{1}{|\textbf{E}|}\sum_{\mathbf{t_1},\mathbf{t_2} \in \textbf{E}} -\omega * \log\left(\frac{\theta_{\mathbf{t_1}, \mathbf{t_2}}}{\pi}\right)|w_{\mathbf{t_1}} - w_{\mathbf{t_2}}|
\end{equation}
where \textbf{E} is the set of mesh edges $\textbf{E} \subset \textbf{F}\times\textbf{F}$, $\theta_{\mathbf{t_1}, \mathbf{t_2}}$ is the dihedral angle between triangles $t_1$,$t_2$ which share an edge, and $\omega$ is an energy scaling term. In all our experiments, we set $\omega = 0.1$. 

Lastly, LSCM does not perform as well with disconnected domains. To guarantee the segmentation boundary encloses a single connected region, we add a \emph{floodfill prior} during training, in which we identify all triangles with probabilites $\geq 0.5$, and take the largest contiguous patch of these triangles which contains the source triangle $\mathbf{t^*}$. We mask out the weights of the other triangles, generating the floodfill weights $w^*$. We pass the floodfill weights into the parameterization layer and pass the resulting UV into our threshold loss. 
In order to discourage spurious weights outside of the compact segmentation boundary, we penalize the original probabilities by passing them into our smoothness loss $L^{\text{smooth}}(w)$.  
Incorporating all of these terms, our final loss function during end-to-end training is
\begin{equation}\label{eq:loss}
    L = L^{\text{threshold}}(w^*) + L^{\text{smooth}}(w)
\end{equation} 
%

\subsection{Near-Developable Segmentation Pretraining}
\label{subsec:synth_data}
In practice we need to initialize our system to produce reasonable segmentations which can be further improved through our end-to-end training. To this end, we pretrain our network to segment developable (flattenable with no distortion) patches with boundaries at sharp creases.

As there is no existing dataset of meshes with labeled distortion-aware decompositions,  we build a synthetic one. We consider a collection of geometry primitives -- cones, cubes, cylinders, spheres, and tetrahedrons, with known developable decompositions (excluding the sphere). To generate training shapes we sample a subset of primitives, apply additional non-rigid deformations (twist, bend, taper, stretch), and compute a CSG union on these perturbed primitives to obtain a watertight mesh. For each mesh, we generate an additional augmented mesh by sampling 10\% of the vertices and jittering them along their normals. We apply Laplacian smoothing to the resulting mesh. See the supplemental for more details and pseudocode. 

For each of our geometry primitives, we take their natural developable decomposition as the starting ground truth (i.e. cubes are segmented into planes, cylinders into the body and two caps, etc.). To create ground truth segmentation labels we retain correspondences to the original primitives, and use them to segment the perturbed mesh. Note, however, that our augmentations do not preserve developability, though developable segmentation is typically too strong of a constraint for practical applications outside of fabrication. Instead we ensure \emph{near-developability} of target segmentations by parameterizing each augmented segmentation with SLIM \cite{rabinovich_scalable_2017}, and discarding all patches with isometric or conformal distortion above a threshold of 0.05. Our measures of isometric and conformal distortion are $D_I = \max(\sigma_1, 1/\sigma_2)$ and $D_C = (\sigma_1 - \sigma_2)^2$, respectively, where $\sigma_1$ and $\sigma_2$ are the singular values of the Jacobian of the UV mapping in decreasing order. 

\begin{figure}[t]
    \centering
    \includegraphics[width=\columnwidth]{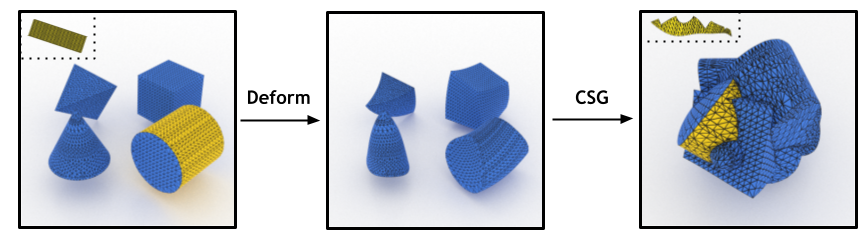}
    \vspace{-.4cm}
    \caption{Generation of our ``near-developable" synthetic dataset: we draw random geometric primitives, deform them, and apply a CSG union to obtain our final synthetic shape, while transferring the correspondence to the original segmentation (yellow). UVs of the corresponding segmentations are shown in the insets.}
    \label{fig:syntheticdataset}
\end{figure}

To generate training samples we simulate user selection by randomly sampling 5\% of the triangles (up to a maximum of 20) on valid near-developable patches (excluding regions corresponding to spheres), to use as initial selections $\mathbf{t^*}$. The dataset construction procedure is illustrated in \cref{fig:syntheticdataset}. Our training dataset consist of 87 meshes with 7,912 initial selections, and our test set consists of 17 meshes with 1,532 initial selections. We pretrain our edge encoder network $\phi_E$ and classifier network $\phi_C$ with L2 supervision using Adam with learning rate $10^{-3}$ for 150 epochs.

\subsection{Distortion Self-Supervised Training}
\label{subsec:training}
In the second phase of training, we train the system end-to-end with the differentiable parameterization layer over a dataset of natural shapes taken from the Thingi10k dataset \cite{zhou_thingi10k_2016}. We download a subset of meshes from Thingi10k, filtering for meshes which are between $3K$ and $20K$ faces, are manifold, and have one connected component. We remesh the shapes using the isotropic explicit remeshing tool from Meshlab~\cite{cignoni_meshlab_2008} so that the meshes have between $5K$ and $12K$ edges. For each remeshed object, we sample $1$\% of the surface triangles to use as training anchor samples, up to a maximum of $50$ faces. Our resulting training set consists of 315 meshes with 6,300 initial selections, and a validation set containing 86 meshes with 1,720 initial selections. 

Similar to the pretraining phase, we batch input pairs of meshes and initial triangle selections. To avoid distribution shift, we additionally draw training samples from the synthetic dataset 50\% of the time and apply strong supervision to those predictions. We perform end-to-end training using Adam with learning rate $10^{-3}$ for 100 epochs.
\subsection{Inference and Patch Postprocessing}\label{subsec:inference}
During inference, we compute UV maps with SLIM~\cite{rabinovich_scalable_2017} instead of LSCM, as the former directly minimizes isometric distortion. Thus the isometric distortion of the LSCM UV map is an \emph{upper bound} of the isometric distortion achieved by the SLIM UV map. We avoid using SLIM in the parameterization layer during training as it is an iterative and heavy optimization technique. However, we consider this to be interesting follow-up work.  As SLIM assumes a disk-topology patch, we run floodfill as before (\cref{subsec:objectives}) to get a single connected component, and apply a seam-cutting algorithm by iteratively choosing the longest boundary loop and its nearest boundary loop, and cutting between them along the shortest path until the patch is disk topology. 

Lastly, in order to straighten the segmentation's boundary and reduce jagged features, we run the graphcuts algorithm~\cite{kolmogorov_what_2004} over our model predictions, where the graph's node potentials are set to $W$ and the edge weights are set to $-\log(\frac{\theta}{\pi})$ where $\theta$ is the dihedral angle between two faces.
\section{Experiments}
\label{sec:exp}
In this section we quantitatively and qualitatively evaluate the distortion-aware selections produced by our method and a few baselines. Since some methods only focus on segmentation heuristics, we post-process all segmentations using SLIM~\cite{rabinovich_scalable_2017} to get the resulting UV parameterizations. 

We describe our evaluation metrics and datasets (\cref{sec:expsetup}), discuss segmentations produced with our method (\cref{sec:ourresults}), compare to baseline alternatives (\cref{sec:baselines}), and run ablations (\cref{sec:ablations}). 



\subsection{Experimental Setup}\label{sec:expsetup}

\noindent\textbf{Datasets.} We evaluate our method and baselines on three datasets. First, we use our Near-Developable Segmentation Dataset, as it is the only dataset that has ground truth parameterization-aware segmentations. We evaluate on a held-out test set composed of 47 meshes and 1,532 (face, segmentation) samples. Second, we create a dataset of natural shapes from Thingi10k~\cite{zhou_thingi10k_2016} models, filtered to have medium face count, manifoldness and a single connected component, pre-processed as described in~\cref{subsec:training}, leading to 110 test shapes with 2,125 samples (in addition to the 401 shapes used during training). These models are mostly originally designed for 3D printing, and cover a wide range of geometric and topological features associated with real objects. Third we use the Mesh Parameterization Benchmark dataset \cite{shay_dataset_2022} composed of 337 3D models originally designed for digital art. We filter for meshes in the same way as with Thingi10k, which leaves us with 32 models and 269 sampled points. While this data is annotated with ground truth artist UV segmentations, global parameterization is a different objective from distortion-aware local segmentation (as discussed in \cref{sec:relatedwork}), so we reserve the comparison to the supplemental.\\\\
\noindent\textbf{Evaluation Metrics.} Our work is motivated by applications that require maximizing patch area, while keeping distortion under some user-prescribed bound. Our Near-Developable Segmentation Dataset is the only dataset with relevant labels, for which we report standard metrics -- accuracy, mean average precision and F1 score. 

When ground truth segmentations are not available, we propose a novel metric \%$D^\lambda_I$, which we define as the percentage of the faces parameterized with isometric distortion below threshold $\lambda$: $D^\lambda_I = |\{t : D_I(t) < \lambda\}|/|\textbf{F}|$,
where the numerator enumerates faces $t$ with distortion below $\lambda$, normalized by the total faces in \textbf{F}. We use the following energy for our isometry measure $D_I = (\max(\sigma_1, \frac{1}{\sigma_2}) - 1)^2$, where $(\sigma_1, \sigma_2)$ are the singular values of the Jacobian of the UV parameterization~\cite{sorkine_bounded-distortion_2002}. We set $\lambda=0.05$, as we find that this distortion level leads to no significant visual distortion. A higher percentage \%$D^\lambda_I$ indicates a larger, more distortion-aware segmentation. We report \%$D^\lambda_I$ for varying $\lambda$ in the supplemental to show that our conclusions are robust.

\subsection{DA Wand Results}
\label{sec:ourresults}

We run our method on all three datasets and show examples in \cref{fig:gallery}. Our method handles meshes with complex geometric features and topological variations, consistently obtaining a large patch which can be parameterized with little distortion. We handle near-developable surfaces of mechanical parts by splitting them into correct primitives such as cylinders, planes, and annuli, even in presence of high frequency noise (A3, B1, B3, C1, C4). Global analysis enabled by MeshCNN allows our method to extend segments far beyond the selection (A1, B2, B4, C3). Our method also discovers large low-distortion patches on organic shapes with no apparent decomposition into primitives (A2, A4, C2), thanks to distortion-guided training.




\subsection{Baselines}
\label{sec:baselines}
\begin{table}[h]
\vspace{-0.1cm}
\centering
\begin{tabularx}{\columnwidth}{lXXX} 
\toprule
& Acc $\uparrow$ & mAP $\uparrow$ & F1 $\uparrow$  \\
\toprule
LogMap & 0.54 & 0.19 & 0.24 \\
DCharts & 0.85 & 0.37 & 0.47\\
DA Wand (Ours) & \textbf{0.91} & \textbf{0.8} & \textbf{0.82} \\
\bottomrule
\end{tabularx}
\vspace{-3mm}
\caption{Accuracy (\textbf{Acc}), Mean average precision (\textbf{mAP}), and F1 score (\textbf{F1}) reported for each respective method on the ground truth segmentations in our synthetic test dataset.}
\label{tab:gtsynthetic}
\vspace{-0.2cm}
\end{table}


\begin{table}[t]
\vspace{-0.1cm}
\centering
\setlength\tabcolsep{3pt}
\resizebox{\columnwidth}{!}{%
\begin{tabular}{lccccccc} 
\toprule
& \multicolumn{2}{c}{Synthetic} & \multicolumn{2}{c}{Thingi10k}  & \multicolumn{2}{c}{Mesh Param.} & Time (s) \\
\toprule
& \%$D_I$ $\uparrow$ & N & \%$D_I$ $\uparrow$  & N & \%$D_I$ $\uparrow$ & N & $t_{seg}\downarrow$ \\ 
\toprule 
LogMap & 6.5 & 236 & 7.5 & 266 & 5.0 & 179 & 0.15 \\
DCharts & 14.3 & 676 & 15.1 & 770 & 1.6 & 2754 & 32.92 \\
DA Wand (Ours) & \textbf{19.9} & 951 & \textbf{19.9} & 1031 & \textbf{14.3} & 1281 & \textbf{0.11} \\
\bottomrule
\end{tabular}}
\caption{Median $\% D_I$ and $N$ reported for each method across the three benchmark datasets. We also report the median time to segmentation ($t_{seg}$) in seconds across all our test samples.}
\label{tab:distortion}
\vspace{-0.2cm}
\end{table}

Though none of the prior methods discussed in \cref{sec:relatedwork} directly address the distortion-aware selection problem, we identify two methods that can be adapted to our problem. 

First is the quasi-developable segmentation technique, DCharts, originally devised for global parameterization-aware segmentation~\cite{julius_d-charts_2005}. The original method uses Lloyd iterations to optimize over multiple charts, which we adapt to single-chart segmentation conditioned on an initial triangle. Starting with the selected triangle, we greedily add triangles with the smallest DCharts energy, defined at each triangle with respect to our patch as the product of the fitting (developability), compactness, and straight boundary energies. As in the original work, we use the same weights ($\alpha=1.0, \beta=0.7, \gamma=0.5$), and in every iteration grow out the patch by adding adjacent triangles with fitting error under the threshold $F_{max} = 0.2$. We maintain a priority queue of triangles adjacent to the chart boundary based on the DCharts energy, and recompute the fitting energy after popping a triangle from the queue to ensure it meets the threshold for the current patch. Following the original work, after each iteration (when the queue becomes empty) we re-estimate the developability proxy vector $N_C$ and angle $\theta$, which are used for the fitting energy, and re-initialize the queue. We terminate once there are no more valid triangles even after re-estimating the developability proxy. 

Second, we adapt a local parameterization technique to guide segmentation. Specifically, we run a version of the logarithmic map implemented with the Vector Heat Method ~\cite{sharp_vector_2019-1} which computes a local parameterization of the whole mesh centered at the selected triangle. We mark all triangles that are parameterized with isometric distortion $D_I$ below 0.05 (our evaluation threshold) as valid, and use the largest contiguous patch of valid triangles containing the selection as the segmentation.

For consistency, we postprocess all baseline results with the graphcuts \cite{kolmogorov_what_2004} and floodfill algorithms as in~\cref{subsec:inference}. The resulting segmentation is guaranteed to have a disk topology, which we parameterize using SLIM~\cite{rabinovich_scalable_2017}.

\begin{figure}[t]
    \centering
    \newcommand{\pl}{3.8}
    \newcommand{\pll}{-3.8}
    \begin{overpic}[width=\columnwidth]{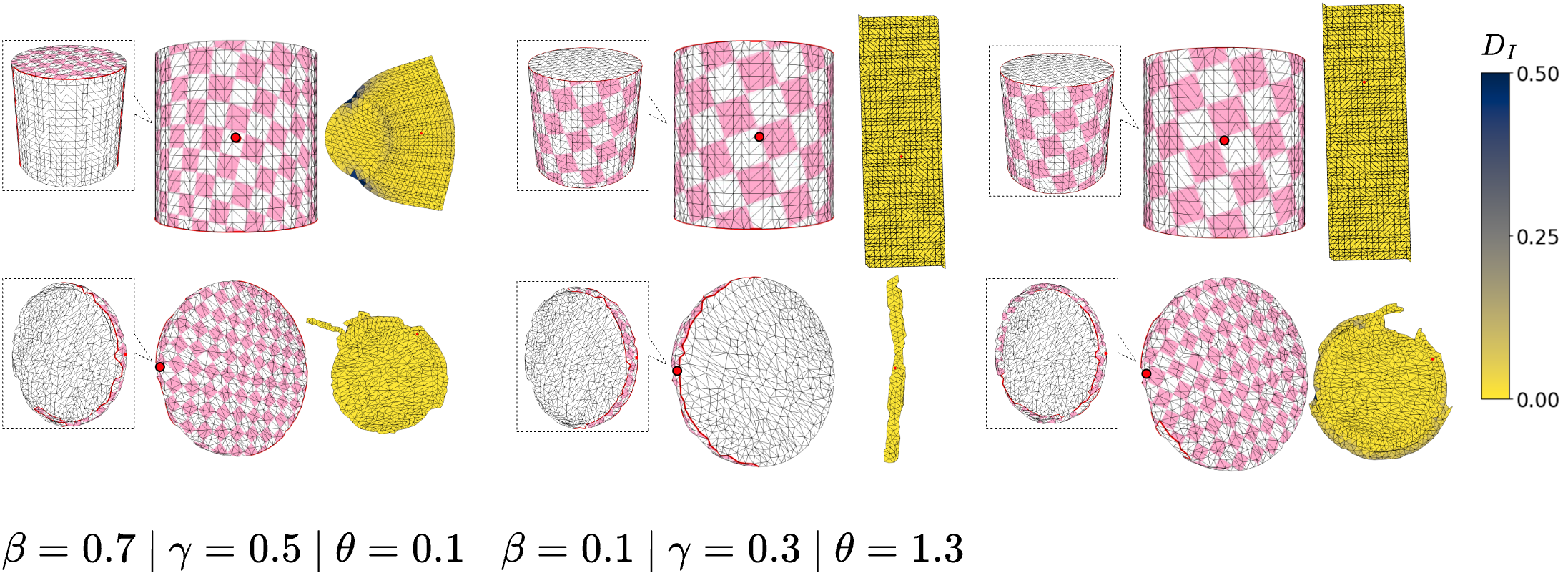}
     \put(3,\pl){\footnotesize\textcolor{black}{\textbf{Default DCharts}}}
     \put(32,\pl){\footnotesize\textcolor{black}{\textbf{Proxy-Biased DCharts}}}
     \put(70,\pl){\footnotesize\textcolor{black}{\textbf{Our Method}}}
    \end{overpic}
    \vspace{-6.5mm}
    \caption{Segmentations predicted by DCharts with default parameters (\textbf{Default DCharts}), DCharts biased towards the developability proxy energy (\textbf{Proxy-Biased DCharts}), and DA Wand (\textbf{Our Method}). The parameter values are given for each DCharts configuration. We show results on a cylinder mesh (top row), and an ashtray mesh from our Thingi10k dataset (bottom row). UV embeddings on the right are colored by isometric distortion $D_I$.}
    \label{fig:dcharts_setting}
\end{figure}

While D-Charts leverages similar developability and compactness priors to our work, these priors are not context-aware, as they only incorporate local patch and triangle information, and the relative importance of these priors have to be manually tuned. We illustrate this in \cref{fig:dcharts_setting}, where the parameters need to be significantly altered for DCharts to produce the expected segmentation on a simple cylinder body. However, these same parameter settings swings the bias in the other direction, so that it segments only the curved region in the ashtray mesh from Thingi10k (second row) while ignoring the large planar patch right next to it. By virtue of its learned inductive bias over many types of surfaces, our method can identify both segmentations without any additional parameter tuning or training.
\rl{SUPP TODO: SHOW SENSITIVITY TO NOISE (bumpy cylinder) + HOW DIFFERENT COST THRESHOLDS INFLUENCE THE SEGMENTATION} \vk{We should probably also mention that their developability prior is very sensitive to high frequency noise (should we show an example of that as well?). We should probably also show how "termination criteria" affects the type of charts we're getting.}

\begin{figure}[t]
    \vspace{-4mm}
    \centering
    \newcommand{\pl}{-2}
    \begin{overpic}[width=\columnwidth]{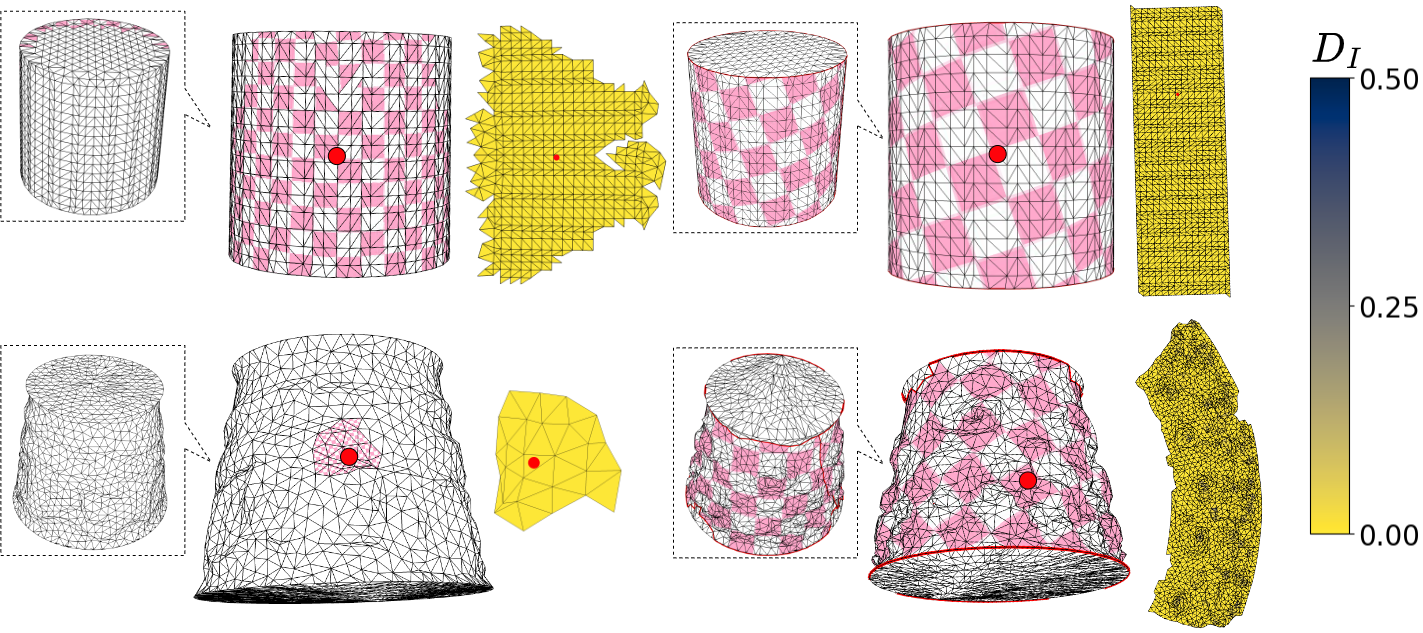}
    \put(10,\pl){\small\textcolor{black}{\textbf{Logarithmic Map}}}
    \put(59,\pl){\small\textcolor{black}{\textbf{Our Method}}}
    \end{overpic}
    \caption{Segmentations predicted on a cylinder mesh (top row) and a noisy cylinder from our Thingi10k dataset (bottom row), for the Logarithmic Map baseline and our method. UV embeddings on the right are colored by isometric distortion $D_I$.}
    \label{fig:expmaps_pathology}
\end{figure}

Our other baseline derives segmentations with the logarithmic map, which makes them trivially distortion-aware. A key limitation, however, is that the logarithmic map parameterizes the entire surface, so cannot ignore regions that introduce heavy distortion (e.g., high-curvature geometry near the selection). We illustrate this in \cref{fig:expmaps_pathology}, where on a simple cylinder, the Exponential Maps baseline segmentation creeps into the top cap, which introduces enough distortion to prevent the segmentation from reaching the far end of the cylinder body. On the other hand, our method identifies the ideal cylindrical segmentation on both the simple cylinder and the noisy sample from the Thingi10k dataset. 

\cref{tab:gtsynthetic} reports the classification performance of each method with respect to the ground truth segmentations for our synthetic test set. Unsurprisingly, our method outperforms the baselines on the segmentation metrics. We report our thresholded-distortion metric \%$D_I^{0.05}$ on all three datasets in \cref{tab:distortion}. We also report the number of segmented triangles (N) and time to segmentation ($t_\text{seg}$) in seconds. Our method outperforms the baselines by a wide margin on all three datasets. Our segmentation time is 25\% faster than the logarithmic maps baseline (LogMap) and several orders of magnitude faster than DCharts. Please see the supplemental for additional qualitative comparisons and time analysis. 

\subsection{Ablations}
\label{sec:ablations}

\begin{table}[t]
\vspace{-0.1cm}
\centering
\setlength\tabcolsep{3pt}
\resizebox{\columnwidth}{!}{%
\begin{tabular}{lcccccc} 
\toprule
& \multicolumn{2}{c}{Synthetic} & \multicolumn{2}{c}{Thingi10k}  & \multicolumn{2}{c}{Mesh Param.} \\
\toprule
& \%$D_I\uparrow$ & N & \%$D_I\uparrow$ & N & \%$D_I\uparrow$ & N \\ 
\toprule 
Our Method (a) & \textbf{19.9} & 951 & \textbf{19.9} & 1031 & \textbf{14.3} & 1281 \\
No floodfill (b) & 15.4 & 1056 & 16.5 & 1113 & 12.3 & 997 \\
No smoothness loss (c) & 10.1 & 1405 & 11.6 & 1500 & 6.8 & 1220 \\
No distortion training (d) & 10.1 & 478.5 & 10.0 & 647 & 3.5 & 647 \\
No mixed training (e) & 4.5 & 2125 & 10.2 & 1729.2 & 2.3 & 886 \\
No pre-training (f) & 4.3 & 2561 & 7.0 & 2537 & 1.2 & 1512 \\
\bottomrule
\end{tabular}}
\vspace{-2mm}
\caption{Median \%$D_I$ and $N$ for our method and different ablation settings across our three benchmark datasets.}
\label{tab:ablation}
\vspace{-0.2cm}
\end{table}

We justify our design choices through a series of ablations (\cref{tab:ablation}). 
Training our method without the compactness priors (b, c) leads to larger but higher distortion patches. Note that the pre-training segmentations (d) are nearly half the size of our final segmentations. On the other hand, removing the pre-training step (f) or not incorporating synthetic dataset samples during distortion self-supervision training (e) significantly degrades the results, which illustrates the importance of this initialization step. 

\vspace{-2mm}
\section{Discussion and Future Work}
We presented a framework which enables training a neural network to learn a data-driven prior for distortion-aware mesh segmentation. Key to our method is a \emph{probability-guided} mesh parameterization layer, which is differentiable, fast, and theoretically grounded. Our wLSCM formulation can be viewed as a way to \emph{relax} the problem of segmenting a local domain which induces a low distortion parameterization. Our experiments demonstrate that our method is an effective technique for producing distortion-aware local segmentations, conditioned on a user-selected triangle.  


Our method holds a few limitations. First, the use of MeshCNN~\cite{hanocka_meshcnn_2019} leads to dependence on the triangulation, in the form of a limited receptive field. Second, the output of our method is not guaranteed to be disk topology (necessary to produce a valid UV map), and is post-processed using a simple cutting algorithm to achieve this property. 

We see many future directions for our work. Instead of using distortion as guidance, other differentiable quantities, such as curvature or visual losses, can be used to guide our segmentation through similar custom differentiable layers. Additionally, we wish to extend our work to predict a segmentation through edges instead of faces, which would enable seam generation within the segmented region.

\vspace{-2mm}
\section{Acknowledgements}
This work was supported in part by gifts from Adobe Research. We are grateful for the AI cluster resources, services, and staff expertise at the University of Chicago. We thank Nicholas Sharp for his assistance with the Polyscope demo and Logarithmic Map code, Georgia Shay for her assistance with the parameterization benchmark, and Vincent Lagrassa for his research assistance. We thank Haochen Wang and the members of 3DL for their helpful comments, suggestions, and insightful discussions.

{\small
\bibliographystyle{ieee_fullname}
\bibliography{main}
}


\newcommand{\beginsupplementary}{%
    \setcounter{section}{0}
	\renewcommand{\thesection}{A\arabic{section}}
	\renewcommand{\thesubsection}{\thesection.\arabic{subsection}}

	\renewcommand{\thetable}{A\arabic{table}}%
	\setcounter{table}{0}

	\renewcommand{\thefigure}{A\arabic{figure}}%
	\setcounter{figure}{0}
}

\beginsupplementary
\twocolumn[{%
 \centering
 {\bf \Large Supplementary Materials for: \\ \ourmethod: Distortion-Aware Selection using Neural Mesh Parameterization}
 \vspace{2em}
}]

\section{Synthetic Dataset Construction}
\label{supp:syntheticdataset}
This section contains more details on how we construct our Near-Developable Dataset. For a given sampled subset of shape primitives (from cube, cone, cylinder tetrahedron, sphere with replacement) normalized to the unit sphere, we loop through each shape primitive and sample a random number of SimpleDeform operations from Blender \cite{blender_online_community_blender_nodate}. The four types of deformations available are [TWIST, BEND, TAPER, STRETCH]. For each deformation operation, we sample a random angle between -0.5 and 0.5 and a random factor quantity from -0.8 to 0.8. Following deformations, we also sample a random axis through the origin and angle and apply a rotation. Finally, we apply a random translation by sampling between -0.3 and 0.3 for each axis. Following the CSG union operation using PyMesh, we apply a 3D augmentation to the resulting shape which involves sampling 1/3 of the vertices of the shape, sliding them by up to 5\% in either direction along their normal, and applying a Laplacian Smoothing operation. 

We maintain correspondences between the ground truth segmentations of each primitive and the final CSG shape. We use this correspondence to map each ground truth segmentation to a segmentation of the CSG shape. There is no guarantee this mapped segmentation is contiguous due to the CSG, so we identify all connected components of the segmentation. We parameterize each connected component separately using SLIM and compute the resulting isometric and conformal distortion $D_I$, $D_C$. If both $D_I$ and $D_C$ are under our threshold of 0.05, then we sample faces from within the segmentation and save them along with the new ground truth segmentation label to generate our labelled data. The full algorithm is shown in \cref{alg:dataset}.

\begin{algorithm}
\caption{Near-Developable Shape Generation}\label{alg:dataset}
\begin{algorithmic}
\Procedure{deform}{$M$, $n$} 
    \For{$i$ in range($n$)}
        \State $\theta \gets \text{Unif}(-0.5, 0.5)$
        \State $\alpha \gets \text{Unif}(-0.8, 0.8)$ 
        \State method = randomChoice(['TWIST', 'BEND', 'TAPER', 'STRETCH']) 
        \State $M \gets \text{blenderSimpleDeform}(\text{method}, \theta, \alpha)$
    \EndFor
\EndProcedure 

\Procedure{randomRotation}{$M$} 
    \State $v \gets \text{Unif}(0,1, \text{size} = 3)$
    \State $\theta \gets \text{Unif}(0, 2\pi)$ 
    \State rotate($M$, $v$, $\theta$) \Comment{rotate $m$ about axis $v$ by $\theta$}
\EndProcedure

\Procedure{randomTranslation}{$M$} 
    \State $M.\text{vertices} \gets \text{Unif}(-0.3, 0.3, \text{size} = 3)$
\EndProcedure

\Procedure{augment}{$M$} 
    \State $vlen \gets$ len(M.vertices)
    \State $vi \gets \text{randomChoice(range(vlen), size=vlen/3})$
    \State $M.\text{vertices}[vi] \gets  M.\text{vertices}[vi] + M.\text{vertexnormals}[vi] * \text{Unif}(0, 0.05, \text{size=length(vi)})$
    \State LaplacianSmooth($M$)
\EndProcedure

\Procedure{generateShape}{$\text{primitives}, \text{segs}$}
\For{$M$ in primitives}
    \State deform($M$, randomChoice(range(3,10)))
    \State randomRotation($M$) 
    \State randomTranslation($M$)
\EndFor 
\State $M', \text{correspondences} \gets \text{csgUnion}(\text{primitives})$
\State $M' \gets \text{augment}(M')$ 
\For{$M$, mmap, seg in zip(primitives, correspondences, segs)}
    \State mappedseg $\gets$ mmap(seg) 
    \State mappedseg $\gets$ collect all connected mapped segmentation components 
    \For{mseg in mappedseg}
        \State $uv \gets\text{SLIM}(\text{mseg})$ 
        \State $ss \gets\text{getSingularValues}(uv)$ 
        \State $d_I \gets \text{mean}((\text{max}(ss[:,0], 1/ss[:,1]) - 1)^2)$
        \State $d_C \gets \text{mean}((ss[:,0] - ss[:,1]))^2)$
        \If{$d_I \leq 0.05$ and $d_C \leq 0.05$}
            \State Sample initial selection faces $f$ in mseg, and save inputs ($M'$, $f$) and corresponding ground truth label mseg 
        \EndIf
    \EndFor
\EndFor
\EndProcedure
\end{algorithmic}
\end{algorithm}

\section{Proof of Theorem 1}
\newtheorem{thm}{Theorem}[section]
\renewcommand{\thethm}{\arabic{thm}}
\begin{thm}
Let $\seg\subset\F$ be a subset of triangles of the mesh, which comprises one connected component. Let $W$ be  non-negative weights assigned to the triangles s.t. the weights of $\seg$ are non-zero. Let $\U_W$ be the minimizer of \cref{e:wlscm} w.r.t $W$. Then $\U_W$, restricted to $\seg$, is a well-defined, continuous function of $W$. Furthermore, if the non-zero weights of $W$ are all equal to $1$, then $\U_W$ restricted to $\seg$ is exactly equal to the (non-weighted) LSCM parameterization of $\seg$.
\end{thm}

\textit{Proof.} LSCM's minimizer is uniquely defined up to a global scaling and rotation which can be chosen by fixing two vertices. Thus, WLOG, we pin two vertices of $\bar{\F}$ to two corners of the unit grid. Then, the LSCM minimization problem over $\bar{\F}$ has a unique solution, as proved in \cite{levy_least_2002}. Assume we satisfy the theorem's assumption and all the non-zero weights of $W$ are equal to 1 and are within $\bar{\F}$. Then, for any parameterization $\U$ and its restriction to $\seg$, denoted $\U|_\seg$, it holds that \begin{align*}
    E_{\text{wLSCM}}(\U) = \sum_{\tri\in\F} w_{\tri}||A_{\tri} - S(A_{\tri})||^2 \\= \sum_{\tri\in\bar{\F}} ||A_{\tri} - S(A_{\tri})||^2 = E_{\text{LSCM}}(\U|_\seg), 
\end{align*} 
namely any minimizer $\U_W$ of $E_\text{wLSCM}$ w.r.t the weights $W$, restricted to $\seg$, is the unique minimizer of LSCM of $\seg$. 

Furthermore, $\U_W|_\seg$ is a minimizer of a strictly-convex quadratic energy \cref{e:wlscm}, and hence is the solution of a linear equation, i.e., $\U_W$ is the root of a linear polynomial with coefficients which are linear combinations of $W$. Since $W$ is non-zero for $\bar{\F}$, $\U_W$ restricted to $\bar{\F}$ is well-defined. Since polynomial roots are continuous in their coefficients, $\U_W$ restricted to $\bar{\F}$ is a well-defined, continuous function of $W$. \hfill $\Box$

\section{$\%D_I^\lambda$ with Changing $\lambda$ and Cost Thresholds}

\begin{figure}[t]
    \centering
    \includegraphics[width=\columnwidth]{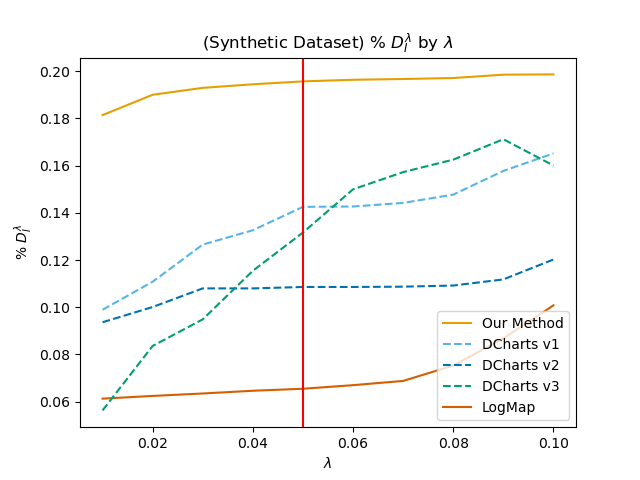}
    \includegraphics[width=\columnwidth]{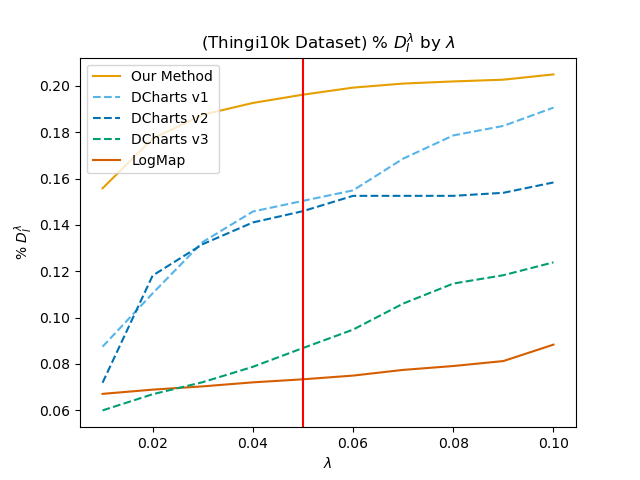}
    \includegraphics[width=\columnwidth]{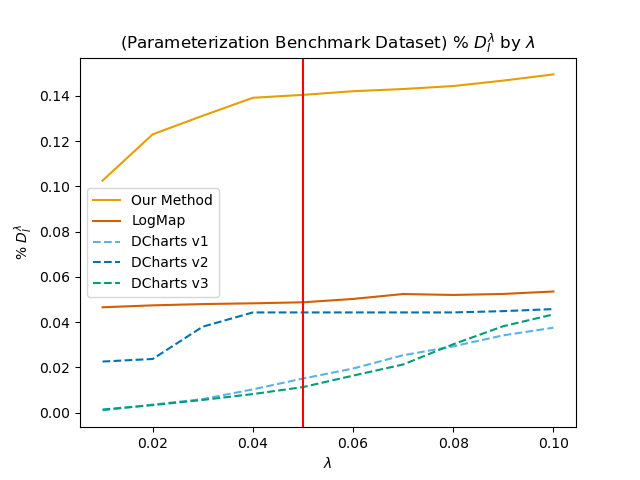}
    \caption{\%$D_I$ metric with sliding $\lambda$ values. ``DCharts v1" refers to DCharts with the default parameters from the main paper.``DCharts v2" and ``DCharts v3'' refer to versions of DCharts with the cost threshold set to 0.1 and 0.3, respectively. The red line marks $\lambda=0.05$, which is the threshold we report in the main paper.}
    \label{fig:lambdas}
\end{figure}

We compare results on the metric \% $D_I^\lambda$ on all 3 datasets for values of $\lambda$ from 0.01 to 0.1 in \cref{fig:lambdas}. ``DCharts v1" refers to DCharts with the default parameters from the main paper.``DCharts v2" and ``DCharts v3'' refer to versions of DCharts with the cost threshold set to 0.1 and 0.3, respectively. For the LogMap baseline, we set the distortion threshold cutoff used for the segmentation heuristic to be equal to $\lambda$. We mark $\lambda=0.05$ with a red line, which is the threshold we report in our main paper.

\section{Postprocessing, Selection Stability, and Timing}
\begin{figure}
    \centering
    \includegraphics[width=\columnwidth]{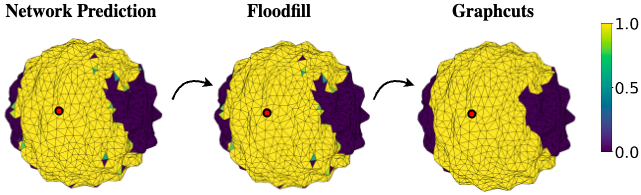}
    \caption{An example of our method's raw prediction and the subsequent effects of the post-processing steps.}
    \label{fig:postprocess}
\end{figure}

\ourmethod~produces compact segmentation results which are disk topology and near-disk topology. In order to guarantee a disk topology segmentation, we apply a floodfill procedure which takes the largest contiguous subset of the segmented region starting from the selection point. We follow up with a graphcuts procedure to smooth out potential jagged edges along the segmentation boundary, which is a standard procedure in mesh segmentation. We visualize these steps in \cref{fig:postprocess}.

\begin{figure}
    \centering
    \includegraphics[width=\columnwidth]{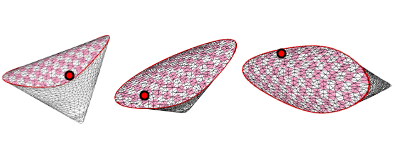}
    \caption{\ourmethod~is highly stable on selections near sharp feature boundaries which border a developable patch.}
    \label{fig:stable}
\end{figure}

\ourmethod~also produces highly stable selections on models with sharp feature curves and clear developable regions. Note how in \cref{fig:stable}, even when the selection point is on the boundary of the feature curve, our method is able to robustly segment the same developable patch bounded by the feature curve. 

\begin{figure}
    \centering
    \includegraphics[width=\columnwidth]{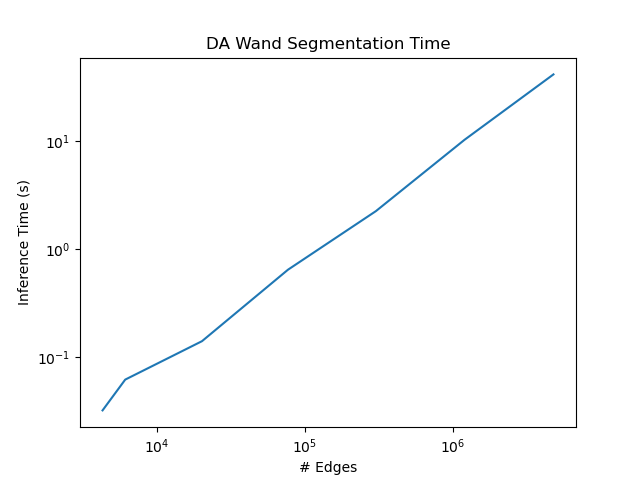}
    \caption{Our method produces segmentations in roughly linear time with respect to the \# edges of the model.}
    \label{fig:segtime}
\end{figure}

Our network architecture builds on MeshCNN, which is a fully convolutional U-Net, implying O(n) time complexity with respect to the input size, in this case the number of edges of the mesh. We show this roughly linear complexity in \cref{fig:segtime}, where we are able to segment models with 1 million edges in around 10 seconds on a RTX2080 gpu. 

\section{Additional Qualitative Comparisons}
\begin{figure*}[t]
    \centering
    \includegraphics[width=\textwidth]{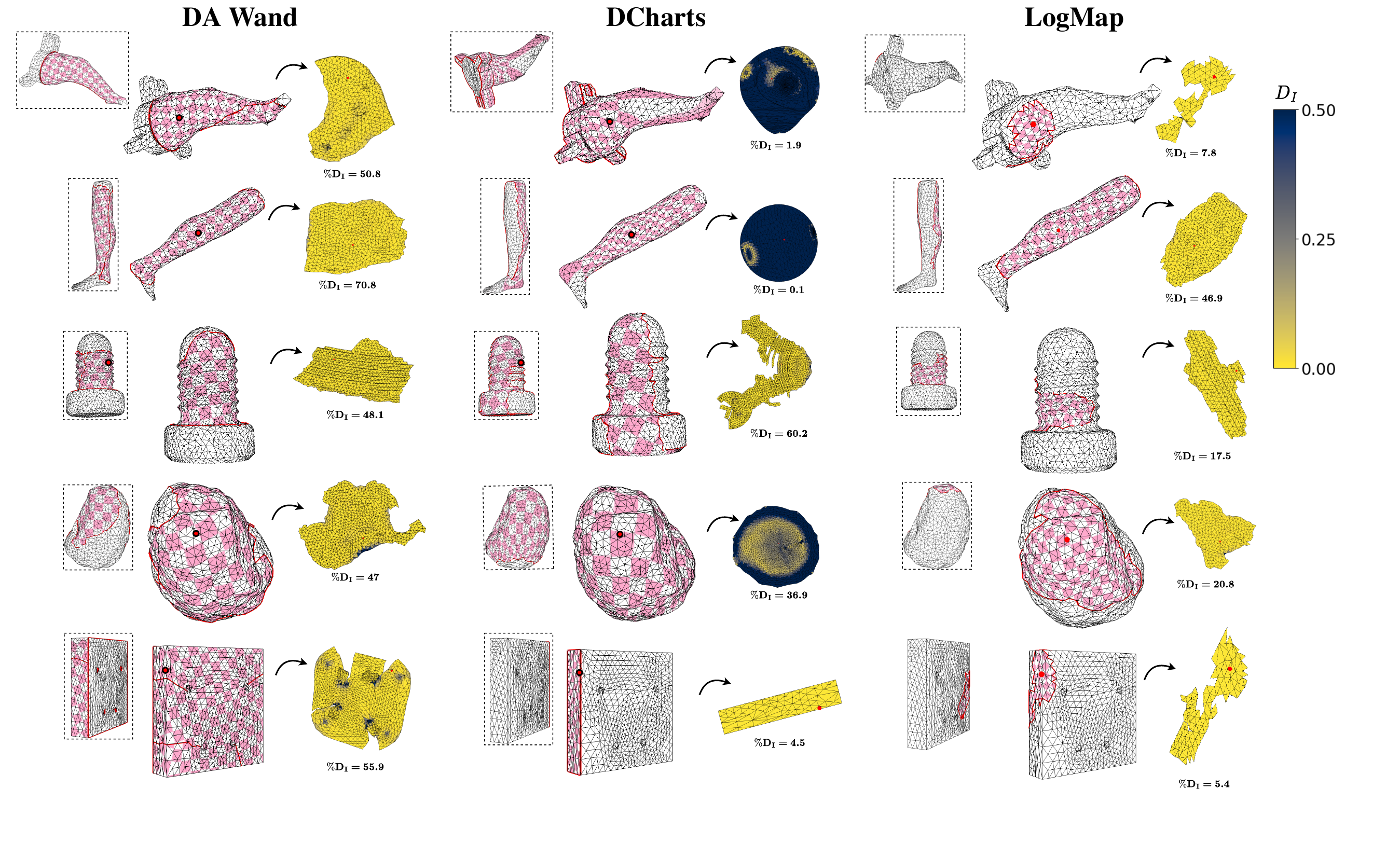}
    \vspace{-10mm}
    \caption{Sample segmentations between \ourmethod, DCharts, and the LogMap baselines.}
    \label{fig:gallerycomparison}
\end{figure*}

We show qualitative comparisons between \ourmethod~and the LogMap and DCharts baseline methods in \cref{fig:gallerycomparison}. Due to the strict distortion threshold and the logarithmic maps' sensitivity to noisy geometry, the LogMap segmentations are low distortion but conservative. On the other hand, DCharts will generally produce large segmentations, but is highly unreliable on natural shapes or shapes with noisy geometry. 

\begin{figure*}[t]
    \centering
    \includegraphics[width=\textwidth]{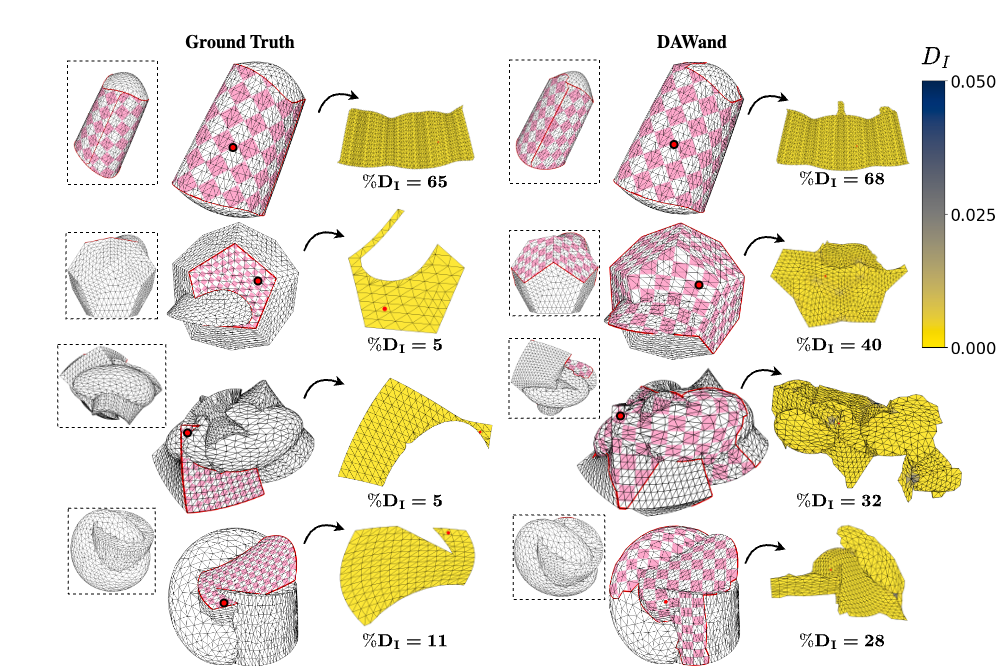}
    \caption{Sample segmentations between \ourmethod~and the ground truth over the synthetic test set. We report the percentage of triangles under isometric distortion 0.05 $\% D_I$ in bold.}
    \label{fig:syntheticcomparison}
\end{figure*}

We also show qualitative comparisons of the ground truth and DA Wand predictions over the synthetic dataset in \cref{fig:syntheticcomparison}. Note that in most cases, the ground truth predictions are constrained to the smallest bounding plane or cylinder of the selection point, whereas our method can segment far beyond nearby feature curves to achieve much larger local parameterizations with little to no cost in distortion.

\section{Parameterization Benchmark: Artist Global Segmentations}

\begin{figure*}[t]
    \centering
    \includegraphics[width=\textwidth]{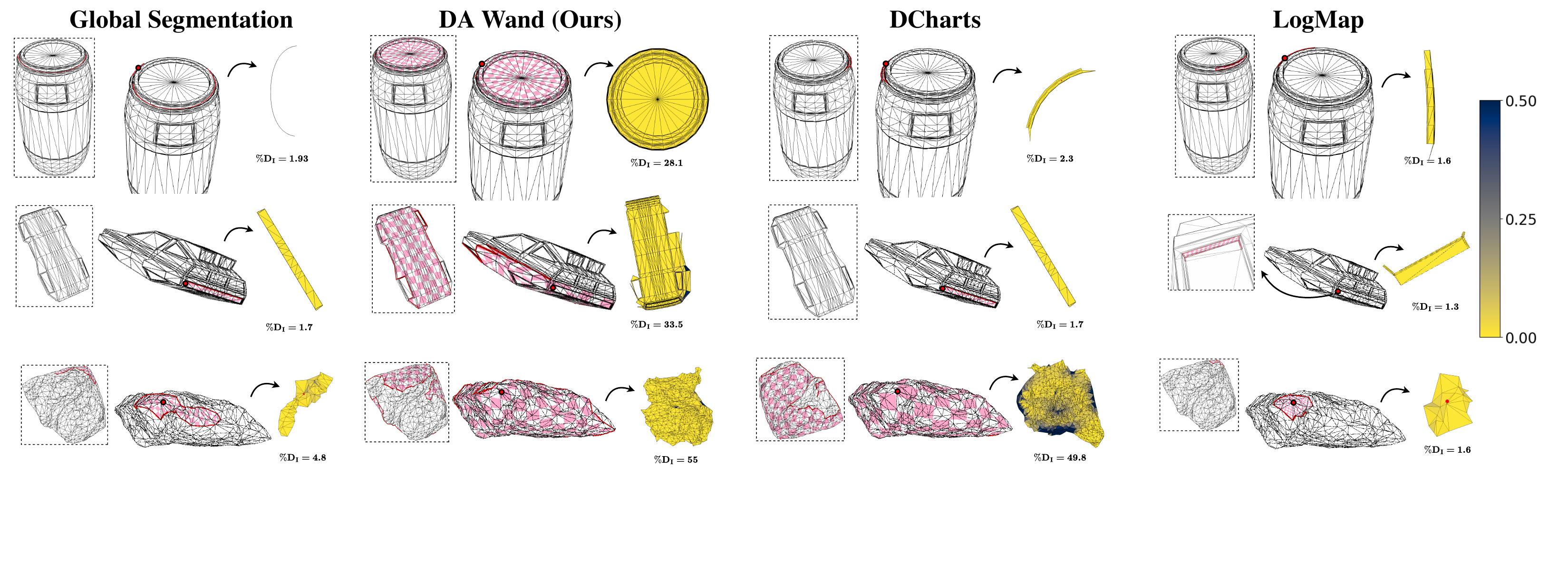}
    \vspace{-15mm}
    \caption{Sample segmentations between the Parameterization Benchmark labels (Global Segmentation), \ourmethod, DCharts, and the LogMap baselines.}
    \label{fig:gallerypbenchmark}
\end{figure*}

We show a few examples of the artist global segmentations which contain our selection points from the Parameterization Benchmark Dataset in \cref{fig:gallerypbenchmark}. These segmentations were intended for \emph{global UV parameterization}, which involves different priors from \emph{local parameterization}, as explained in \cref{sec:relatedwork}. From \cref{fig:gallerypbenchmark} it is clear that our segmentations are preferable in the context of local texturing or decaling, as we achieve large segmentations with little to no tradeoff in terms of parameterization distortion.

\end{document}